\documentclass{article}

\usepackage{PRIMEarxiv}

\usepackage{amssymb}
\usepackage{amsmath}
\usepackage{graphicx}
\usepackage[flushleft]{threeparttable}
\usepackage{tabularx}
\usepackage{url}

\title{DynaSeg: A Deep Dynamic Fusion Method for Unsupervised Image Segmentation Incorporating Feature Similarity and Spatial Continuity}

\author{
  Boujemaa Guermazi\\
  Electrical, Computer, and Biomedical Engineering \\
  Toronto Metropolitan Unversity \\
  \texttt{bguermazi@torontomu.ca} \\
   \And
  Riadh Ksantini \\
  Computer Science \\
  University of Bahrain \\
  \texttt{rksantini@uob.edu.bh} \\
  \And
  Naimul Khan\\
   Electrical, Computer, and Biomedical Engineering \\
  Toronto Metropolitan Unversity \\
  \texttt{n77khan@torontomu.ca} \\}
  
\begin{document}
  %\begin{frontmatter}
  \maketitle
%% Abstract
\begin{abstract}
Our work tackles the fundamental challenge of image segmentation in computer vision, which is crucial for diverse applications. While supervised methods demonstrate proficiency, their reliance on extensive pixel-level annotations limits scalability. We introduce DynaSeg, an innovative unsupervised image segmentation approach that overcomes the challenge of balancing feature similarity and spatial continuity without relying on extensive hyperparameter tuning. 
Unlike traditional methods, DynaSeg employs a dynamic weighting scheme that automates parameter tuning, adapts flexibly to image characteristics, and facilitates easy integration with other segmentation networks. By incorporating a Silhouette Score Phase, DynaSeg prevents undersegmentation failures where the number of predicted clusters might converge to one. DynaSeg uses CNN-based and pre-trained ResNet feature extraction, making it computationally efficient and more straightforward than other complex models. Experimental results showcase state-of-the-art performance, achieving a 12.2\% and 14.12\% mIOU improvement over current unsupervised segmentation approaches on COCO-All and COCO-Stuff datasets, respectively. We provide qualitative and quantitative results on five benchmark datasets, demonstrating the efficacy of the proposed approach. Code available at  Code available at \url{https://github.com/RyersonMultimediaLab/DynaSeg}
\end{abstract}

%% Keywords

%\end{frontmatter}

\section{Introduction}
\label{Intro}
In recent years, computer vision has taken remarkable strides, driven by the availability of large-scale image datasets as well as the development of advanced machine learning algorithms. Within various real-time applications \cite{medsurvey}\cite{zhao2023lif}\cite{Ruban2019}\cite{kaymak2019}, image segmentation stands out as a pivotal computer vision task, playing a crucial role in interpreting visual content at a granular level. Unlike the straightforward task of image classification, which allocates a category label to the entire image \cite{sanchez2013image}, segmentation involves assigning category labels to individual pixels, effectively outlining the image into meaningful segments or regions. This nuanced approach not only enhances the interpretation of complex scenes but also contributes to a deeper understanding of visual content. 

The segmentation task can be approached from different perspectives, leading to three main types of image segmentation. Semantic segmentation \cite{gao2022fbsnet} identifies uncountable and shapeless regions, often referred to as 'stuff,' such as grass, sky, or road, based on similar textures or materials. Instance segmentation \cite{hafiz2020survey} focuses on pinpointing and segmenting individual instances of countable objects, such as people, animals, and tools, treating them as distinct 'things.' Lastly, panoptic segmentation \cite{kirillov2019panoptic} unifies the two distinct concepts used to segment images, assigning a semantic label to each pixel in an image and giving the pixel a unique identifier if it is an object instance. The semantic nuances that differentiate these tasks have led to the creation of methods that employ specialized architectures. 
% However, variability in object appearance, complex interactions between objects, variations in object scale, and noisy or ambiguous edges present significant challenges in image segmentation. 
However, despite the advancements in segmentation methodologies, significant challenges persist. Variability in object appearance, complex interactions between objects, variations in object scale, and noisy or ambiguous edges all pose obstacles to accurate segmentation. Objects can look different due to lighting, colour, and texture changes, making consistent identification difficult. Complex interactions, such as overlapping people in a crowd or intertwined tree branches and power lines, further complicate segmentation. Additionally, objects appear in various sizes depending on their distance from the camera, requiring algorithms to handle different scales effectively. Noisy or ambiguous edges, caused by low resolution or motion blur, add another layer of difficulty.
Developing robust and adaptable segmentation methods that can address these complexities is crucial for improving the accuracy and reliability of image segmentation in real-world applications. Therefore, while task-specific architectures have significantly advanced segmentation objectives, there is a pressing need for methods that offer more flexibility to generalize across different segmentation tasks, ultimately enhancing the performance of segmentation algorithms in challenging real-world scenarios.

Classical pioneering segmentation techniques such as active contour models (ACM) \cite{ACM}, k-means \cite{Kmeans}, and graph-based segmentation (GS) \cite{graph}\cite{liu2012unsupervised} impose global and local data and geometry constraints on the masks. As a result, these techniques are sensitive to initialization and require heuristics such as point resampling, making them unsuitable for modern applications. 
Currently, the state of the art in segmentation is dominated by deep learning.  The Mask R-CNN framework \cite{MRCNN} has provided advancement in semantic and instance image segmentation. However, a major drawback is a substantial requirement for hand-labelled data, limiting the widespread applicability across diverse domains. This challenge becomes particularly pronounced in the context of pixel-wise classification, where the cost of annotation per image is prohibitively expensive.

Unsupervised image segmentation is a possible solution to automatically segment an image into semantically similar regions where the system can find objects, precise boundaries and materials that even the annotation system may not unveil properly. The task has been studied as a clustering problem in the recent literature \cite{infomax}\cite{DIC}\cite{IIC}, with promising results.
Differentiable feature clustering, a state-of-the-art CNN-based algorithm proposed by \cite{them}, simultaneously optimizes pixel labels and feature representations through a combination of \textit{feature similarity} and \textit{spatial continuity} constraints \cite{them}. Feature similarity corresponds to the constraint that pixels in the same cluster should be similar to each other. Spatial continuity refers to the constraint that pixels in the same cluster should be next to each other (continuous). However, to achieve the desired segmentation result, \cite{them} applies a manual parameter tuning to find the optimal balancing weight $\mu$, which fails to achieve a good balance between the two constraints, as mentioned above, depending on the degree of detail in the image and the dataset.

In this work, we present a novel dynamically weighted loss scheme, offering flexibility in updating the parameters and automatically tuning the balancing weight $\mu$. Our approach conditions the value of $\mu$ based on the number of predicted clusters and the iteration number. At each iteration, we dynamically prioritize one of the constraints, resulting in a well-balanced optimization. Key contributions include \footnote{A preliminary version of this work was published in \cite{guermazi}.}:

\begin{itemize}
  \item \textbf{Dynamically Weighted Loss:} Introducing a flexible and adaptive loss function that dynamically adjusts the balancing weight $\mu$ during training, ensuring optimal balance between feature similarity and spatial continuity. This makes the model adaptable to different datasets without requiring manual changes to parameters, such as the balancing weight, and facilitates easy implementation of the proposed method on other segmentation networks.
  \item \textbf{Silhouette Score Phase:}  Incorporating a silhouette score-based phase to guide the optimization process, ensuring improved cluster quality. Unlike traditional methods that rely on fixed thresholds to determine segmentation termination and may converge to a single cluster, failing to capture the diversity and complexity of real-world images, our approach dynamically assesses cluster quality using silhouette scores. This prevents under-segmentation failures, where the segmentation process prematurely stops, leading to inaccurate results. By dynamically evaluating cluster quality, our method ensures that the segmentation process continues until meaningful clusters are formed, without the need for predefined thresholds. This not only improves segmentation accuracy but also reduces the burden of fine-tuning additional hyperparameters.
  % Incorporating a silhouette score-based phase to guide the optimization process, ensuring improved cluster quality. This prevents under-segmentation failures, where relying solely on feature similarity and spatial continuity with fixed weight factors can lead to the number of predicted clusters diverging to one.
  \item \textbf{ResNet with FPN Integration:} Leveraging the power of ResNet combined with the Feature Pyramid Network (FPN) for enhanced feature extraction in the unsupervised segmentation task. The FPN enhances the representation map, making it semantically strong and improving the overall segmentation performance. This integration is particularly beneficial for segmenting objects at different scales and resolutions, as FPN enables multi-scale processing and feature integration across different levels of abstraction. Additionally, by incorporating ResNet, we capitalize on its inherent strengths, similar to those of CNNs, while leveraging residual information to mitigate issues like vanishing gradients. This combination amplifies the network's capacity to capture intricate details and semantic information critical for accurate segmentation.
\end{itemize}

Experimental results across five benchmark datasets demonstrate that our method, DynaSeg, achieves superior quantitative metrics and qualitative segmentation results. The dynamically weighted loss facilitates a more effective balance between feature similarity and spatial continuity, leading to enhanced segmentation performance.

\section{Related Work}
Recent advancements in machine learning have significantly impacted various vision applications, particularly image segmentation. The introduction of convolutional neural networks (CNNs) and the availability of large annotated datasets have revolutionized image segmentation, enabling these models to surpass traditional algorithms such as active contour models (ACM) \cite{ACM}, k-means clustering \cite{Kmeans}, and graph-based segmentation (GS) \cite{graph}. Notable CNN-based techniques, including Mask R-CNN \cite{MRCNN}, U-Net \cite{Unet}, SegNet \cite{badrinarayanan2017segnet}, and DeepLab \cite{chen2017deeplab}, offer precise segmentation capabilities. However, these models are heavily reliant on large annotated datasets, which not only escalate annotation costs but also introduce challenges related to missing labels and data quality variations \cite{ADE20K, lecun2010mnist, krizhevsky2009learning}. These limitations have driven the development of unsupervised segmentation models. Among these approaches, clustering-based and CNN-based methods have gained prominence, each offering unique strategies to address these challenges. Our model, DynaSeg, exemplifies this trend by providing scalable and cost-efficient solutions without the need for explicit annotations.

% These limitations have driven the development of unsupervised segmentation models, including our model, DynaSeg, which provides scalable and cost-efficient solutions without the need for explicit annotations.

% \subsection{Unsupervised Segmentation}
% Recent advancements in unsupervised segmentation methods offer a cost-effective alternative, aiming to discover objects, boundaries, and materials without explicit ground truth. Among these approaches, clustering-based and CNN-based methods have gained prominence, each offering unique strategies to tackle the challenges of unsupervised segmentation.

\subsection{Clustering-Based Methods}
% Foundational unsupervised segmentation methods like K-means \cite{Kmeans}, Mean Shift \cite{meanShift}, and graph-based segmentation (GS) \cite{graph} form clusters based on pixel feature similarity but often lack spatial coherence. These methods treat each pixel independently, resulting in fragmented segmentations with disconnected regions, especially images with complex structures and textures. In contrast, our model incorporates both feature similarity and spatial continuity to ensure that similar and contiguous pixels belong to the same cluster. Unlike static clustering methods \cite{Kmeans}\cite{meanShift} that require the number of clusters to be specified a priori, DynaSeg adjusts the number of clusters during training, enhancing the model's flexibility and robustness.

Foundational unsupervised segmentation methods like K-means \cite{Kmeans}, Mean Shift \cite{meanShift}, and graph-based segmentation (GS) \cite{graph} cluster pixels based on feature similarity but often lack spatial coherence. These methods treat each pixel independently, resulting in fragmented segmentations, especially in images with complex structures. In contrast, our model integrates feature similarity and spatial continuity, ensuring that similar and contiguous pixels form the same cluster. Unlike static methods requiring a predetermined number of clusters \cite{Kmeans}\cite{meanShift}, DynaSeg adjusts the number of clusters during training, enhancing flexibility and robustness.

% One significant clustering-based approach is Deep Image Clustering (DIC) \cite{DIC}, which breaks down the segmentation problem into feature transformation and trainable deep clustering sub-networks. However, it relies on superpixels, introducing predetermined boundaries. Similarly, Invariant Information Clustering (IIC) \cite{IIC}, and its versions \cite{mutualInf} uses mutual-information-maximization for clustering. However, its reliance on enforcing a uniform distribution over clusters makes it particularly effective with well-balanced datasets \cite{lecun2010mnist, krizhevsky2009learning}. Yet, the efficacy of IIC \cite{IIC} is contingent upon data balance and can be influenced by the specific data augmentation strategy employed \cite{Autoregressive, augment, SegSort}. 

A notable approach is Deep Image Clustering (DIC) \cite{DIC}, which transforms features and clusters them using deep sub-networks but relies on superpixels with predetermined boundaries. Similarly, Invariant Information Clustering (IIC) \cite{IIC} and its variants \cite{mutualInf} use mutual-information-maximization but require uniform cluster distributions, making them effective with balanced datasets \cite{lecun2010mnist, krizhevsky2009learning}. Their efficacy can be influenced by data augmentation strategies \cite{Autoregressive, augment, SegSort}.
Unlike methods that separate feature extraction and clustering \cite{DIC, IIC}, DynaSeg employs Joint Optimization where the backpropagation of clustering losses directly influences the CNN. This integration ensures a cohesive segmentation process, continuously refining the model and overcoming the limitations of predetermined boundaries and data imbalance.
% Unlike the methods \cite{DIC}\cite{IIC} that separate feature extraction and clustering into distinct steps, DynaSeg employs Joint Optimization where the backpropagation of clustering losses directly influences the CNN. This integration ensures a cohesive segmentation process that is continuously refined, overcoming the limitations of predetermined boundaries and data imbalance.

\subsection{Deep Learning-Based Methods}

% Deep learning-based methods have garnered significant attention in unsupervised image segmentation, showcasing remarkable capabilities across various approaches. For instance, Generative Adversarial Networks (GANs) \cite{generative} \cite{BigBigGAN} \cite{Labels4Free} \cite{GAN2} have emerged as powerful tools for creating segmented images by effectively separating foreground and background. Specifically, the GAN-based approach proposed by Melas-Kyriazi \cite{generative} leverages pre-trained GANs like BigBiGAN \cite{BigBigGAN} to extract salient object segmentation from the latent space. Despite their innovation, these methods often rely on some form of class supervision, which inherently limits their purely unsupervised nature. Labels4Free \cite{Labels4Free} extends Style-GAN2 \cite{GAN2} with a segmentation branch for unsupervised foreground object segmentation, typically focusing on binary clustering. However, these approaches tend to rely on discriminative losses, such as binary cross-entropy loss, for training the discriminator network, which may not directly optimize for segmentation quality, resulting in suboptimal outcomes.

Deep learning-based methods have gained significant attention in unsupervised image segmentation, showcasing various innovative approaches. Generative Adversarial Networks (GANs) \cite{generative} \cite{BigBigGAN} \cite{Labels4Free} \cite{GAN2} have emerged as powerful tools for creating segmented images by separating foreground and background. For example, Melas-Kyriazi \cite{generative} leverages pre-trained GANs like BigBiGAN \cite{BigBigGAN} to extract salient object segmentation from the latent space. Despite their innovation, these methods often require some form of class supervision, limiting their purely unsupervised nature. Labels4Free \cite{Labels4Free} extends Style-GAN2 \cite{GAN2} with a segmentation branch for unsupervised foreground object segmentation, typically focusing on binary clustering. However, these approaches tend to rely on discriminative losses, such as binary cross-entropy loss, for training the discriminator network, which may not directly optimize segmentation quality, resulting in suboptimal outcomes.
In contrast, our model adopts a hybrid approach for multi-class segmentation by integrating a clustering strategy with CNN-based feature extraction.

Beyond generative models, contrastive learning \cite{Picie2021}\cite{DenseSiam} has gained traction for its ability to learn representations by contrasting similar and dissimilar pairs of image patches. PICIE \cite{Picie2021} introduces a Siamese perspective to unsupervised segmentation, emphasizing equivariance learning and consistent clustering assignments across different image views. DenseSiam \cite{DenseSiam} focuses on learning dense representations through contrastive learning and clustering-based techniques. While these methods effectively capture images' inherent structures and patterns, contributing significantly to segmentation performance, they have notable limitations. Specifically, these approaches introduce additional computational operations, struggle with unknown transformations, and are sensitive to noisy or low-quality data, often resulting in biased representations or suboptimal clustering results.
Our model addresses these limitations and offers a more robust solution. DynaSeg employs a spatial continuity loss that acts as a high-pass filter, enhancing segmentation accuracy by preserving high-frequency details and attenuating low-frequency components. This discourages smooth regions from being segmented into multiple clusters, promoting homogeneity within each cluster, making the segmentation more coherent and reducing noise within clusters.

Wnet \cite{Wnet} combines U-net \cite{Unet} structures into auto-encoders for segmentation, followed by a post-processing phase for refinement. While effective, this method is computationally expensive and requires extensive hyperparameter tuning. Infomax \cite{infomax} presents an alternative where an input image is partitioned into superpixels. The Region-Wise-Embedding (RWE) extracts a feature embedding for each superpixel region, and Mutual-Information-Maximization is employed with adversarial training. Despite its efficiency, it faces challenges like superpixel reliance and predetermined boundaries.
In contrast, our proposed hybrid model offers a streamlined solution by eliminating the need for post-processing. Our model generates compact clusters directly from the feature space by dynamically optimizing the clustering loss and spatial continuity loss. This integrated approach enables our model to extract rich and informative features, facilitating accurate and robust segmentation, even in challenging scenarios. DynaSeg offers enhanced flexibility and adaptability by removing the reliance on predefined boundaries or superpixels \cite{old}, ensuring more natural and visually appealing segmentation outcomes.
\begin{figure*}[htb]

\hfill
\begin{minipage}[b]{1.0\linewidth} 
  \centering
  \centerline{\includegraphics[width=18cm]{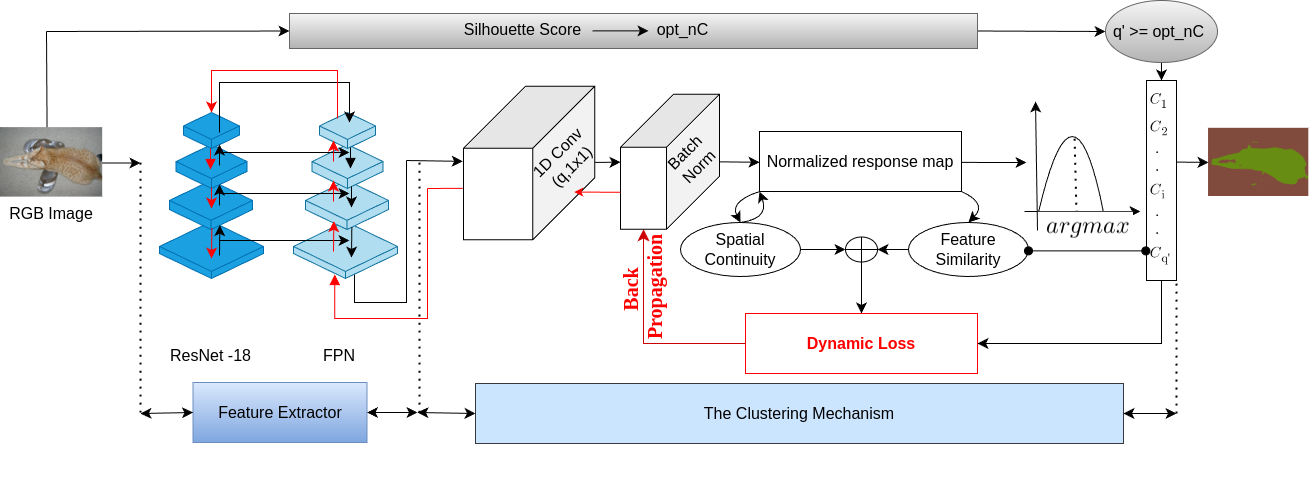}}
%  \vspace{1.5cm}
\end{minipage}
\caption{Overview of the DynaSeg Framework: The Feature Extractor Network generates a feature map, classified into $q$ clusters via a linear classifier and batch normalization, resulting in a normalized response map. Cluster labels $c_i$ are assigned to each pixel using the argmax function. The number of clusters $q'$ is dynamically updated based on feature similarity and spatial continuity. Loss $L$ is computed during backpropagation, with parameters updated using SGD. This process iterates $T$ times to refine cluster labels $c_i$, achieving final segmentation. The Silhouette Score sets $opt\_nC$ as the threshold for $q'$ to prevent under-segmentation. Black arrows indicate the feedforward path, while red arrows represent backpropagation.}
\label{DynaSegNet}
\end{figure*}
% Differentiable feature clustering, as proposed by \cite{them}, optimizes pixel labels and feature representations using a combination of feature similarity and spatial continuity constraints. However, achieving the desired segmentation result with this approach often requires manual parameter tuning to find the optimal balancing weight ($\mu$) between feature similarity and spatial continuity. This manual tuning process can be cumbersome and may fail to balance the two constraints well, mainly when dealing with images of varying levels of detail and datasets with different characteristics. Moreover, relying solely on fixed weight factors for feature similarity and spatial continuity constraints can lead to the number of predicted clusters converging to one, making it challenging to adapt to different datasets and requiring extensive fine-tuning. Additionally, the lack of adaptability and the need for manual parameter adjustments make integrating this approach with other networks difficult.
% In contrast, our model introduces a Dynamic Weighting Scheme that dynamically adjusts the weighting parameter ($\mu$) during iterations. By incorporating a silhouette score-based phase to guide the optimization process, our model prevents under-segmentation failures and ensures robust performance across diverse datasets. This dynamic approach not only simplifies the integration of our model with other networks but also enhances its adaptability and scalability, making it well-suited for various segmentation tasks and datasets.

Additionally, recent advancements in CNN architectures, such as Vision Transformers (ViT) \cite{du2023mdvit}, have shown promise in various computer vision tasks, including image segmentation. However, ViTs come with their own set of challenges, such as high computational complexity, data inefficiency, training instability, and a propensity for overfitting, which can limit their accessibility and applicability in specific contexts. In contrast, our hybrid model offers a more efficient and scalable solution by dynamically optimizing the clustering loss and leveraging a CNN-based network or a pre-trained ResNet with FPN to extract high-level features from input images.  This approach ensures robustness and adaptability across different datasets. 
% Overall, our model provides a practical and effective solution for unsupervised image segmentation, addressing the limitations of existing methods and presenting promising opportunities for real-world applications.

\subsection{Dynamic Weight Adjustment}
Dynamic loss functions have been explored in several contexts within machine learning. For example, Dynamic Autoencoders (DynAE) \cite{mrabah2020deep} employ adaptive loss functions to balance reconstruction and clustering objectives dynamically during training, enhancing clustering performance by adapting to the data's changing characteristics. Similarly, DFNet \cite{jiang2019dfnet} utilizes dynamic loss weights to manage class imbalance and refine segmentation accuracy in semantic segmentation tasks. By calculating weights based on the pixel count of each class within a batch, DFNet mitigates the imbalance and improves performance, particularly for minority classes.

% The application of dynamic loss functions in image segmentation, particularly in unsupervised settings, has received relatively less attention. Most existing segmentation methods rely on fixed weights that do not adapt during training, which can limit their effectiveness across diverse datasets. DynaSeg, however, introduces a unique dynamic loss function specifically designed for unsupervised image segmentation.
% One of the key benefits of DynaSeg is its ability to autonomously balance competing objectives, thereby enhancing segmentation quality without the need for manual hyperparameter tuning. By integrating a silhouette score-based phase, our model guides the optimization process, preventing under-segmentation failures and ensuring robust performance across diverse datasets. 
The application of dynamic loss functions in image segmentation, particularly in unsupervised settings, has received relatively less attention. Most existing segmentation methods rely on fixed weights, such as Differentiable Feature Clustering \cite{them}, which optimizes pixel labels and feature representations using a combination of feature similarity and spatial continuity constraints. However, achieving the desired segmentation result with this approach often requires manual parameter tuning to find the optimal balancing weight ($\mu$) between feature similarity and spatial continuity. This manual tuning process can be cumbersome and may fail to balance the two constraints well, especially when dealing with images of varying levels of detail and datasets with different characteristics. Moreover, relying solely on fixed weight factors for feature similarity and spatial continuity constraints can lead to the number of predicted clusters converging to one, making it challenging to adapt to different datasets and requiring extensive fine-tuning. Additionally, the lack of adaptability and the need for manual parameter adjustments make integrating this approach with other networks difficult. In contrast, DynaSeg introduces a unique dynamic loss function specifically designed for unsupervised image segmentation. This model employs a Dynamic Weighting Scheme that dynamically adjusts the weighting parameter ($\mu$) during iterations. By incorporating a silhouette score-based phase to guide the optimization process, our model prevents under-segmentation failures and ensures robust performance across diverse datasets. This dynamic approach not only simplifies the integration of our model with other networks but also enhances its adaptability and scalability, making it well-suited for various segmentation tasks and datasets. One of the key benefits of DynaSeg is its ability to autonomously balance competing objectives, thereby enhancing segmentation quality without the need for manual hyperparameter tuning.

% Differentiable feature clustering, as proposed by \cite{them}, is a CNN-based algorithm that optimizes pixel labels and feature representations using a combination of feature similarity and spatial continuity constraints \cite{them}. However, achieving the desired segmentation result with this approach often requires manual parameter tuning to find the optimal balancing weight ($\mu$) between feature similarity and spatial continuity. This manual tuning process can be cumbersome and may fail to balance the two constraints well, mainly when dealing with images of varying levels of detail and datasets with different characteristics. Moreover, relying solely on fixed weight factors for feature similarity and spatial continuity constraints can lead to the number of predicted clusters converging to one, making it challenging to adapt to different datasets and requiring extensive fine-tuning. Additionally, the lack of adaptability and the need for manual parameter adjustments make integrating this approach with other networks difficult.
% In contrast, our model introduces a Dynamic Weighting Scheme that dynamically adjusts the weighting parameter ($\mu$) during iterations. By incorporating a silhouette score-based phase to guide the optimization process, our model prevents under-segmentation failures and ensures robust performance across diverse datasets. This dynamic approach not only simplifies the integration of our model with other networks but also enhances its adaptability and scalability, making it well-suited for various segmentation tasks and datasets.
\section{Methodology}
This section details the proposed DynaSeg method, including the architecture design, dynamic weighting scheme, silhouette score integration, and training procedure. Each component is meticulously described to ensure reproducibility and clarity.
\subsection{Model Architecture}
Our unsupervised image segmentation model, DynaSeg, is designed to effectively capture complex patterns and structures within images. Building upon the architecture introduced in our preliminary work \cite{guermazi}, we present further enhancements tailored to adapt to varying image content. The architecture comprises three main components: a Feature Extractor Network, a Dynamic Weighting Scheme, and a Clustering Mechanism.

As shown in Figure (\ref{DynaSegNet}), the Feature Extractor Network produces a $p$-dimensional feature map $r$, which is then classified into $q$ classes using a linear classifier layer. A batch normalization function is applied to obtain a normalized map $r'$. Lastly, the argmax function assigns each pixel a cluster label $c_n$ based on the maximum value in $r'$. Each pixel is assigned the corresponding cluster label $c_n$, identical to allocating each pixel to the closest point among the $q'$ representative points.

The loss $L$ (defined later) is calculated during backward propagation, and the convolutional filters and classifier parameters are updated using stochastic gradient descent. This process is iterated $T$ times to achieve the final prediction of cluster labels $c_n$. The segmentation is handled in an unsupervised manner, with the number of clusters $q'$ dynamically updated based on feature similarity and spatial continuity.

The components of DynaSeg work synergistically: the feature extractor captures complex details, the dynamic weighting scheme adapts the model's focus during training, and the clustering mechanism assigns meaningful labels to pixels. The integration of these components is crucial for achieving state-of-the-art unsupervised image segmentation.

\subsection{Feature Extractor Network}
The Feature Extractor Network, functioning as the primary processing stage, is tasked with extracting high-level features from input images. This pivotal role lays the foundation for subsequent stages by unravelling intricate details and capturing nuanced patterns within the input images. In our extended model, we explore two alternatives for feature extraction: a CNN-based Network comprising $M$ convolutional components and a pre-trained ResNet \cite{resnet} with a Feature Pyramid Network (FPN) decoder. This exploration allows us to compare and leverage the strengths of each approach for improved image segmentation, ensuring robustness and efficiency in feature extraction.

\subsubsection{CNN-based Network}
Our CNN-based network consists of $M$ convolutional components, each integrating 2D convolution, ReLU activation, and batch normalization. This architecture forms a robust feature extraction pipeline, producing a $p$-dimensional feature map $x_n$. The network's simplicity ensures computational efficiency, making it suitable for practical applications without the need for extensive computational resources.

Pooling layers often reduce the spatial resolution of feature maps, leading to the loss of fine-grained details crucial for accurate image segmentation. By avoiding pooling, our network retains the original resolution of the feature maps, preserving critical spatial information. This design choice enhances the network's ability to produce high-quality, semantically meaningful segments without sacrificing computational efficiency. Additionally, the CNN-based approach's robustness in handling high-dimensional data makes it particularly effective for various image segmentation tasks.

\subsubsection{Pre-trained ResNet with Feature Pyramid Network (FPN) Decoder}
We incorporate a pre-trained ResNet-18 model with a Feature Pyramid Network (FPN) decoder to enhance our network's feature extraction capabilities. While pyramid networks have been utilized in segmentation tasks, such as in APCNet \cite{apcnet} and FPN for Land Segmentation \cite{fpn_land}, our methodology incorporates several novel elements that distinguish DynaSeg from these existing methods. 
Unlike traditional pyramid networks \cite{apcnet} that construct multi-scale contextual representations using global-guided local affinity, DynaSeg extends this by dynamically adjusting the segmentation process, effectively balancing feature similarity and spatial continuity to produce more nuanced and context-aware segmentation outputs. This dynamic adjustment addresses the limitations of static parameter settings in conventional pyramid networks. 
Additionally, our approach uniquely integrates a pre-trained ResNet with an FPN decoder specifically tailored for unsupervised segmentation, enhancing feature extraction through robust multi-resolution processing. optimized for unsupervised learning, distinguishing it from other FPN methods \cite{apcnet} \cite{fpn_land} .

\begin{itemize}
    \item \textbf{ResNet Adaptation:} We modify the ResNet-18 architecture by removing the Global Average Pooling (GAP) and Fully Connected (FC) layers, typically used for classification tasks. Instead, we replace the final fully connected layer with a convolutional layer to produce spatially preserved feature maps suitable for segmentation tasks. ResNet's use of residual connections effectively mitigates the vanishing gradient problem, allowing for deeper network training and more accurate feature extraction.
    \item \textbf{FPN Decoder:} We integrate a Feature Pyramid Network (FPN) decoder instead of employing a standard decoder. The FPN utilizes lateral connections to fuse features from different resolutions, enhancing the representation of multi-scale features. By applying the FPN on top of the high-level features from the ResNet (specifically the Conv5\_x layer), we generate a $p$-dimensional feature map $x_n$. This approach retains high-resolution features that are semantically rich and detailed.
\end{itemize}

This strategic combination leverages ResNet-18's depth and robust feature extraction capabilities along with FPN's multi-scale feature integration. As a result, our model can capture and utilize details at various granularity levels, significantly improving image segmentation performance. The residual connections in ResNet-18 help in mitigating the vanishing gradient problem, allowing for deeper and more accurate feature extraction. This complements the FPN's ability to integrate features at multiple scales, providing a comprehensive representation that enhances segmentation accuracy.

The dual approach of using both a CNN-based network and a pre-trained ResNet with FPN decoder allows us to compare and leverage the strengths of each method. The CNN-based network offers robustness and computational efficiency, making it suitable for scenarios with limited computational resources. In contrast, the pre-trained ResNet with FPN decoder provides richer and more diverse feature representations, which are crucial for capturing fine-grained details in complex images. By incorporating both approaches, we achieve a balanced and highly effective feature extraction mechanism that enhances the overall performance of our segmentation framework.

\subsection{Dynamic Weighting Scheme}
% The Dynamic Weighting Scheme dynamically adjusts the weighting parameter ($\mu$) during iterations, introducing two variants: Feature Similarity Focus (FSF) and Spatial Continuity Focus (SCF). This adaptive scheme enables the model to shift its focus between feature similarity and spatial continuity, reducing sensitivity to parameter tuning and enhancing overall segmentation performance. By employing a dynamically weighted loss, the model becomes adaptable to different datasets without requiring manual changes to parameters, facilitating easy implementation of the proposed method on other segmentation networks.
The Dynamic Weighting Scheme is a novel contribution of our method, designed to dynamically adjust the weighting parameter (\(\mu\)) during training iterations. This adaptive approach offers several key advantages:

\textbf{- Flexibility and Robustness:} Traditional methods often rely on fixed weighting parameters, which may not be optimal for all datasets or image complexities. Our dynamic weighting scheme continuously adapts \(\mu\), ensuring an optimal balance between feature similarity and spatial continuity throughout training. This reduces the need for extensive manual parameter tuning, enhancing the model's robustness and usability across diverse applications.

\textbf{- Improved Performance:} By adjusting \(\mu\) dynamically, the model can emphasize feature similarity or spatial continuity as needed during different training stages. This adaptability leads to improved segmentation performance, as the model fine-tunes itself to the specific characteristics of the data being processed.

The dynamic adjustment of \(\mu\) addresses specific challenges such as the difficulty in maintaining a consistent balance between feature similarity and spatial continuity and the sensitivity of traditional models to fixed parameter settings. By continuously evaluating and updating \(\mu\), our scheme ensures that the model remains adaptable and effective across different training states.

\subsubsection{Loss Function}
Our methodology incorporates specific loss functions to strike a balance between feature similarity and spatial continuity—two pivotal criteria for distinguishing pixel clusters. The feature similarity loss, denoted as $L_{sim}$ (Equation (\ref{lsim})), plays a crucial role in ensuring that pixels with similar features receive identical labels. This loss quantifies the cross-entropy between the normalized response map $r'_n$ and the corresponding cluster labels $c_n$. Minimizing this loss facilitates the extraction of precise attributes, significantly contributing to segmentation accuracy.
\begin{equation}\label{lsim}
    L_{\text{sim}}(r'_n, c_n) = -\sum_{i=1}^{N} \sum_{j=1}^{q} \delta(j - c_i) \ln r'_{i,j},
\end{equation}

where  $r'_n$: normalized response; $c_n$: cluster labels; \( \delta(t) \) is the Kronecker delta function defined as:

\[
\delta(t) = 
\begin{cases} 
1 & \text{if } t = 0, \\
0 & \text{otherwise}.
\end{cases}
\]
Conversely, the spatial continuity loss $L_{con}$ acts as a high-pass filter, ensuring that spatially continuous pixels receive the same label. Defined by Equation (\ref{L1_total}), $L_{con}$ as the Manhattan Distance $L1$ Norm of horizontal and vertical differences in the response map $r'_n$. This mitigates the deficiencies caused by superpixels \cite{old}, efficiently removing excess labels due to complex patterns or textures and ensuring spatially continuous pixels share the same label. This loss also promotes homogeneity within each cluster, enhancing segmentation coherence and reducing noise within clusters.

\begin{equation}\label{horizontal_difference}
    \Delta h_{i,j} = |r'_{n}(i, j) - r'_{n}(i, j+1)|
\end{equation}

\begin{equation}\label{vertical_difference}
    \Delta v_{i,j} = |r'_{n}(i, j) - r'_{n}(i+1, j)|
\end{equation}

\begin{equation}\label{L1_horizontal}
    L1_h(r'_n) = \sum_{i=1}^{H} \sum_{j=1}^{W-1} |\Delta h_{i,j}|
\end{equation}

\begin{equation}\label{L1_vertical}
    L1_v(r'_n) = \sum_{i=1}^{H-1} \sum_{j=1}^{W} |\Delta v_{i,j}|
\end{equation}

\begin{equation}\label{L1_total}
    L_{\text{con}}(r'_n) = L1_h(r'_n) + L1_v(r'_n)
\end{equation}

Equations (\ref{horizontal_difference}) and (\ref{vertical_difference}) quantify the horizontal ($\Delta h_{i,j}$) and vertical differences ($\Delta v_{i,j}$) between adjacent pixels in the response map $r'_n$, effectively capturing changes along the horizontal and vertical axes. The subsequent computations, as defined by Equations (\ref{L1_horizontal}) and (\ref{L1_vertical}), aggregate the absolute horizontal differences and vertical differences, giving rise to measures $L1_h(r'_n)$ and $L1_v(r'_n)$. These measures represent the cumulative horizontal and vertical changes in the response map, respectively.

The culmination of these individual components is expressed in Equation (\ref{L1_total}), where $L_{\text{con}}(r'_n)$ represents the total spatial continuity loss. This holistic measure is derived by summing both horizontal and vertical components. In essence, Equation (\ref{L1_total}) succinctly captures the pixel-wise Manhattan distances along both axes, culminating in the \(L1\) Norm. This comprehensive measure effectively encapsulates spatial continuity by considering both horizontal and vertical proximity, contributing collectively to the overarching continuity criterion.

The unsupervised segmentation loss function is represented by Equation (\ref{oldloss}):

\begin{equation}\label{oldloss}
  L = L_{sim}(\{r'_n,c_n\}) + \mu L_{con}(\{r'_n\}),  
\end{equation}
Here, $\mu$ represents the weight for balancing.
\begin{figure}[htb]

\hfill
\begin{minipage}[b]{1.0\linewidth}
  \centering
  \centerline{\includegraphics[width=9cm]{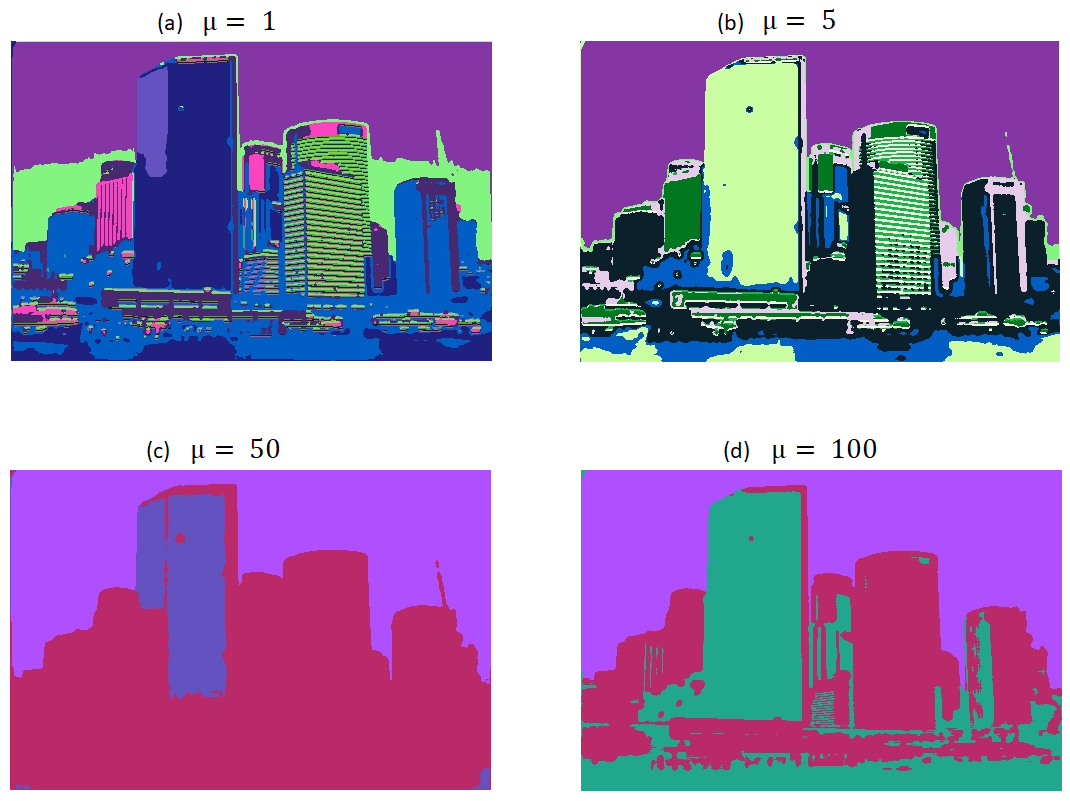}}
%  \vspace{1.5cm}
\end{minipage}
\caption{Results for different $\mu$ values on a sample image from the BSD500 dataset.}
\label{muVariation}
\end{figure}

While the combination of feature similarity ($L_{sim}$) and spatial continuity ($L_{con}$) losses in the unified function $L$ yields reasonably accurate unsupervised segmentation results, the parameter $\mu$ plays a crucial role and can lead to varying outcomes. The sensitivity to $\mu$ is illustrated in Figure (\ref{muVariation}), underscoring the challenge of selecting an appropriate value. As can be seen, for $\mu=50$ and $\mu=100$, the segmentation is coarse, resulting in sky, buildings, and coastal regions. However, the image is further segmented with $\mu= 1$ and $\mu= 5$, where buildings are further segmented into glass buildings, concrete buildings, and different floors. Although the authors in \cite{them} argue that the value of $\mu$ is proportional to the coarseness of segmentation, We see that the results are not consistent, e.g. the segmentation for $\mu=50$ appears coarser than  $\mu=100$. This inconsistency poses a practical problem. The high sensitivity to the parameter means that extensive tuning is required for each dataset to obtain semantically meaningful results.

\begin{table*}[t]
\begin{center}
\begin{threeparttable}
\begin{tabularx}{\textwidth}{|c|X|X|X|X|X|}
\hline
\textbf{} & \multicolumn{5}{|c|}{\textbf{dataset}} \\
\cline{2-6} 
\textbf{Method} & \textbf{\textit{BSD500 All}}& \textbf{\textit{BSD500 Fine}}&  \textbf{\textit{BSD500 Coarse}}&\textbf{\textit{BSD500 Mean}}&\textbf{\textit{PASCAL VOC2012}} \\
\hline
IIC \cite{IIC}  & 0.172 & 0.151 & 0.207 & 0.177 & 0.273 \\ 
\hline
k-means clustering & 0.240 & 0.221 & 0.265 & 0.242 & 0.317 \\ 
\hline
Graph-based Segmentation \cite{graph} & 0.313 & 0.295 & 0.325 & 0.311 & 0.365 \\ 
\hline
CNN-based + superpixels \cite{old}  & 0.226 & 0.169 & 0.324 & 0.240 & 0.308 \\
\hline
CNN-based + weighted loss, $\mu=5$ \cite{them} & 0.305 & 0.259 & 0.374 & 0.313 & 0.352  \\ 
\hline
Self-supervised Multi-view Clustering \cite{fang2021self} & 0.316 & 0.266 & 0.391 & 0.339 & 0.383\\
\hline
Double Clustering with Superpixel Fitting \cite{li2024differentiable} & 0.338 & 0.291 & 0.385 & 0.348 & 0.376 \\
\hline
DynaSeg - Spatial Continuity Focus (SCF)& 0.330 & 0.290 & 0.407 & 0.342 & {\bf0.396} \\ 
\hline
DynaSeg - Feature Similarity Focus (FSF)& {\bf0.349} & {\bf0.307} & {\bf0.420} & {\bf0.359} & 0.391 \\
\hline
\end{tabularx}
\caption{Comparison of $mIOU$ for unsupervised segmentation on BSD500 and PASCAL VOC2012. Best scores are in bold.}
\label{tab1}
\end{threeparttable}
\end{center}
\end{table*}

The competitive nature of both $L_{sim}$ (Feature Similarity Loss) and $L_{con}$ (Spatial Continuity Loss) in our methodology serves to strike a balance between two crucial aspects of unsupervised segmentation:
\begin{itemize}
  \item \textbf{Feature Similarity (Clustering)}: $L_{sim}$ encourages pixels with similar features to be grouped into the same cluster. This is vital for capturing the intrinsic similarities among pixels and achieving meaningful clusters representing distinct objects or regions in the image.
\item \textbf{Spatial Continuity (Smoothness):} Inversely, $L_{con}$ ensures spatially continuous pixels receive the same label. This promotes spatial coherence in the segmentation results and contributes to the smoothness of the segmented regions.
\end{itemize}
The competitiveness arises from the fact that these two objectives may sometimes conflict. For instance, promoting spatial continuity may result in smoother but less accurate segmentation if not balanced properly with the need to capture fine-grained feature similarities. Conversely, focusing too much on feature similarities may lead to fragmented segmentation without considering the spatial arrangement of pixels.
By making $L_{sim}$and $L_{con}$ competitive, we implicitly guide the model to find an optimal trade-off between capturing feature similarities and maintaining spatial coherence. This competitive interaction helps the unsupervised segmentation model produce results that are both accurate in terms of content and visually coherent.

In our innovative approach to unsupervised image segmentation, we introduce a dynamic weighting scheme that addresses these inherent challenges. Recognizing the significance of adaptability during training, our method incorporates a novel dynamic adjustment for the weighting parameter $\mu$. This adjustment responds to the evolving number of predicted clusters and ongoing training iterations, resulting in a flexible and responsive balancing weight.

This dynamic weighting scheme involves changing the weighting parameter's value during training. We observe that prioritizing feature similarity in the earlier training iterations and gradually shifting focus to spatial continuity (or vice versa) enhances adaptability. The proposed approach introduces a dynamic loss function with a continuous variable $\mu$, where the weight $\mu$ dynamically adjusts based on the number of predicted clusters and iterations. We present and compare two versions of this innovative dynamic weighting scheme:

\begin{itemize}

\item \textit{\textbf{Feature Similarity Focus (FSF)}}: Initiate the training process by prioritizing the identification of continuous regions, gradually transitioning to a heightened focus on feature similarity. In this case, the performed trials lead to a linear function of the number of clusters (q') for the new dynamic balancing weight $\mu = (q'/\alpha)$ as shown in equation(\ref{equ3}). We tried other versions that vary exponentially with the value of $q'$; However, such functions resulted in a rapid change in the value of $\mu$, which was not conducive to the balance we sought to achieve between the two constraints. 
\begin{equation}\label{equ3}
    L_{FSF} = L_{sim}(\{r'_n,c_n\}) + (q'/\alpha)   L_{con}(\{r'_n\})
\end{equation}
\item \textit{\textbf{Spatial Continuity Focus (SCF)}}: Initiate training with a focus on feature similarity criteria, gradually transitioning to a stronger emphasis on spatial continuity. In this scenario, the proposed dynamic weight, denoted as $\mu$, takes the form of the multiplicative inverse of the number of clusters, defined as $\mu = (\alpha/q')$ (as shown in Equation (\ref{equ2})). While exploring alternative formulations, including exponential functions, we found that an exponential decay led to an overly rapid change in the value of $\mu$, making it less effective in achieving the desired balance.
\begin{equation}\label{equ2}
    L_{SCF} = L_{sim}(\{r'_n,c_n\}) + (\alpha/q')  L_{con}(\{r'_n\}) 
\end{equation}

\end{itemize}

\subsubsection{Silhouette Score Phase}
We introduce a critical component known as the Silhouette Score Phase to solve an issue in the existing approaches. The clustering model in \cite{them} and \cite{guermazi} only rely on pixel similarity and spatial continuity criteria, assigning the same label to pixels with similar and spatially continuous features. This can lead to a simple solution with q' = 1, causing under-segmentation. By evaluating the compactness and separation of clusters, the Silhouette Score acts as a corrective measure to prevent such under-segmentation issues, ensuring the model achieves a balanced and meaningful segmentation outcome.

The Silhouette Score is a metric widely used in unsupervised learning tasks, particularly clustering, to quantify the goodness of a clustering technique. In the context of our unsupervised image segmentation model, the Silhouette Score Phase operates as follows: at the first iteration of processing the raw image, the Silhouette Score is calculated based on the initial cluster labels. This score serves as a valuable criterion to determine the optimal number of clusters for the given image. By leveraging the Silhouette Score, we introduce a self-regulating mechanism that prevents the model from converging to a single cluster or excessively splitting into numerous small clusters.

\subsection{The Clustering Mechanism}
A linear classifier generates a response map \(r_n = W_c x_n\), initiating the clustering process. This response map undergoes a normalization step, leading to \(r'_n\) with zero mean and unit variance. Employing intra-axis normalization before applying the argmax classification for assigning cluster labels, transforms the original responses \(r_n\) into \(r'_n\) and introduces a preference for a larger $q'$, enhancing the model's adaptability to varying image content. Consequently, the cluster label $c_n$ for each pixel is determined by selecting the dimension with the maximum value in $r'_n$, a process referred to as the argmax classification. This intuitive classification rule aligns with the overarching goal of clustering feature vectors into $q'$ clusters.

This clustering mechanism assigns each pixel to the closest representative point among the $q'$ points, strategically placed at an infinite distance on the respective axes in the $q'$-dimensional space. It is noteworthy that $C_i$ can be $\emptyset$, allowing the number of unique cluster labels to flexibly range from 1 to $q'$.

In summary, the Dynamic Weighting Scheme in DynaSeg significantly enhances segmentation accuracy and flexibility by integrating several innovative approaches. The dynamically weighted loss function allows the model to adapt to different datasets without manual parameter adjustments, simplifying implementation across various segmentation networks. Adaptive clustering adjusts the number of clusters dynamically during training, enhancing the model's flexibility and robustness to accommodate the varying complexities of different datasets. Joint optimization seamlessly integrates feature extraction and clustering, with the backpropagation of clustering losses directly influencing the CNN, overcoming limitations associated with predetermined boundaries and data imbalance. The spatial continuity loss further improves segmentation accuracy by preserving high-frequency details and promoting homogeneity within each cluster, resulting in more coherent segmentations with reduced noise. DynaSeg's design eliminates the need for predefined boundaries or superpixels, providing enhanced flexibility and adaptability, making the segmentation process robust to input data variations and effective across a wide range of scenarios. Overall, these integrated features lead to more accurate, coherent, and adaptable segmentation results.

% our methodology presents a comprehensive approach to unsupervised image segmentation. The interplay of the Feature Extractor Network, Dynamic Weighting Scheme, and Clustering Mechanism, coupled with the innovative Silhouette Score Phase, contributes to state-of-the-art segmentation performance.

\section{Experimental Results}

The objective of our experiments is to showcase the effectiveness of our proposed dynamic weighting scheme for achieving semantically meaningful image segmentation. We conducted extensive evaluations on multiple datasets, comparing against state-of-the-art methods.

\subsection{Experiment Setup}

In our experiments, we introduce two versions of the feature extractor: a CNN-based extractor and a pre-trained ResNet-18 \cite{resnet} with Feature Pyramid Network (FPN). For the CNN-based extractor, we set the number of components in the feature extraction phase, denoted as $M$, to $3$. Conversely, for the ResNet18 with FPN, we modify the output channel to match the number of classes in the dataset (e.g., 27 on COCO stuff-thing dataset).

For consistency across all tests, we set the dimensions of the feature space, $p$, and the cluster space, $q$, to be equal, with both values set to $100$. We employ a learning rate with a base lr=0.1 and SGD optimizer, where the weight decay is 0.0001, and the SGD momentum optimizer is set to 0.9.
The best $\alpha$ for SCF and FSF clustering were experimentally determined from \{25, 45, {\bf50}, 55, 60, 75 100, 200\} and \{2, 10, {\bf15}, 25,50, 100\}, respectively. We report results for $\alpha$ = 15 for FSF clustering ; $\alpha$ = 50 for SCF clustering. 

The mean Intersection Over Union ($mIOU$) is reported for the benchmark datasets. It's important to note that ground truth is utilized solely during the assessment phase and plays no role in the training process.

\subsubsection{COCO-Stuff}
In accordance with the methodology outlined in \cite{IIC, Picie2021, DenseSiam}, we evaluate the performance of our model using the COCO-Stuff dataset  \cite{coco}. This dataset is distinguished for its extensive collection of scene-centric images, featuring 80 thing and 91 stuff categories. The model is evaluated on curated subsets \cite{IIC, Picie2021} of the COCO val2017 split, consisting of 2,175 images,  Following the preprocessing procedure detailed in \cite{Picie2021}, we amalgamate classes to establish 27 categories, comprising 15 stuff and 12 things. It's noteworthy that, unlike earlier studies focusing solely on stuff categories, our evaluation encompasses both things and stuff categories.

\subsubsection{BSD500}
Additionally, we utilize the Berkley Segmentation Dataset BSD500 \cite{BSD} and PASCAL Visual Object Classes 2012 \cite{pascal} for both quantitative and qualitative evaluation of the segmentation results. The BSD500 dataset comprises 500 color and grayscale natural images. Following the experimental setup in \cite{them}, we use the 200 color images from the BSD500 test set to evaluate all models.

\begin{table}[!t]
\renewcommand{\arraystretch}{1.3}
% \extrarowheight

\centering
\begin{tabular}{|l|l|l|l|}
    \hline
    Method & Backbone & mIoU All \\
    \hline
    Modified DC \cite{caron2018deep} & - & 9.8 \\
    IIC \cite{IIC} & ResNet18 & 6.7 \\
    Picie \cite{Picie2021} & ResNet18 & 14.4 \\
    DenseSiam \cite{DenseSiam}& ResNet18 & 16.4 \\
    DynaSeg - FSF & CNN-Based & 30.51 \\
    DynaSeg - SCF & CNN-Based & 27.57 \\
    DynaSeg - FSF & ResNet-18 + FPN & 30.07 \\
    {\bf DynaSeg - SCF} & {\bf ResNet-18 + FPN} & {\bf30.52} \\
    \hline
  \end{tabular}
  \caption{Comparison of mIOU for unsupervised segmentation on COCO-All.}
  \label{tab:coco_all_results}
\end{table}

\begin{table}[!t]
\renewcommand{\arraystretch}{1}
% \extrarowheight
\centering
\begin{tabularx}{0.5\textwidth}{|l|l|X|X|}
    \hline
    Method & Backbone & mIoU Stuff & pAcc \\
    \hline
    IIC \cite{IIC} & ResNet18 & 27.7 & 21.8 \\
    Picie \cite{Picie2021} & ResNet18 & 31.48 & 50.0 \\
    DenseSiam \cite{DenseSiam}& ResNet18 & 24.5 & - \\
    HSG \cite{ke2022unsupervised} & ResNet50 & 23.8 & 57.6 \\
    ReCo+  \cite{shin2022reco} & DeiT-B/8 & 32.6 & 54.1 \\
    STEGO \cite{hamilton2022unsupervised} & ViT-B/16 & 23.7 & 52.5 \\
    DINO + ACSeg \cite{li2023acseg}   & ViT-B/8 & 16.4 & - \\
    DINO + HP \cite{seong2023leveraging}  & ViT-B/8 & 24.6 & 57.2 \\
    DINO + CAUSE-TR \cite{kim2023causal}   & ViT-B/8 & 41.9 & 74.9 \\
    
    DynaSeg - SCF  & CNN-Based & 42.41 & 76.7 \\
    DynaSeg - SCF & ResNet-18 FPN & 42.41 & 76.7 \\
    DynaSeg - FSF & CNN-Based & 42.37 & 76.6 \\
    {\bf DynaSeg - FSF}  & {\bf ResNet-18 FPN} & {\bf54.10} & {\bf81.1} \\
    \hline
  \end{tabularx}
  \caption{Comparison of different methods based on their mean Intersection over Union (mIoU) for 'Stuff' categories and pixel accuracy (pAcc) on COCO-Stuff.}
  \label{tab:coco_stuff_results}
\end{table}

\subsubsection{PASCAL VOC2012}
For PASCAL VOC2012 \cite{pascal}, we treat each segment as an individual entity, disregarding object classification. The VOC2012 dataset is expansive, containing 17,124 images, of which 2,913 have semantic segmentation annotations. We use the 2,913 semantic segmentation images for evaluating our method. In addition, we utilize select images from the Icoseg \cite{Icoseg} and Pixabay \cite{pixab} datasets to present qualitative results. 
% All experimental settings align with those in \cite{them} for a fair and consistent comparison.
\begin{figure*}[!htbp]

%\hfill
%\begin{minipage}[b]{1.0\linewidth} 
  \centering
  \centerline{\includegraphics[width=17cm]{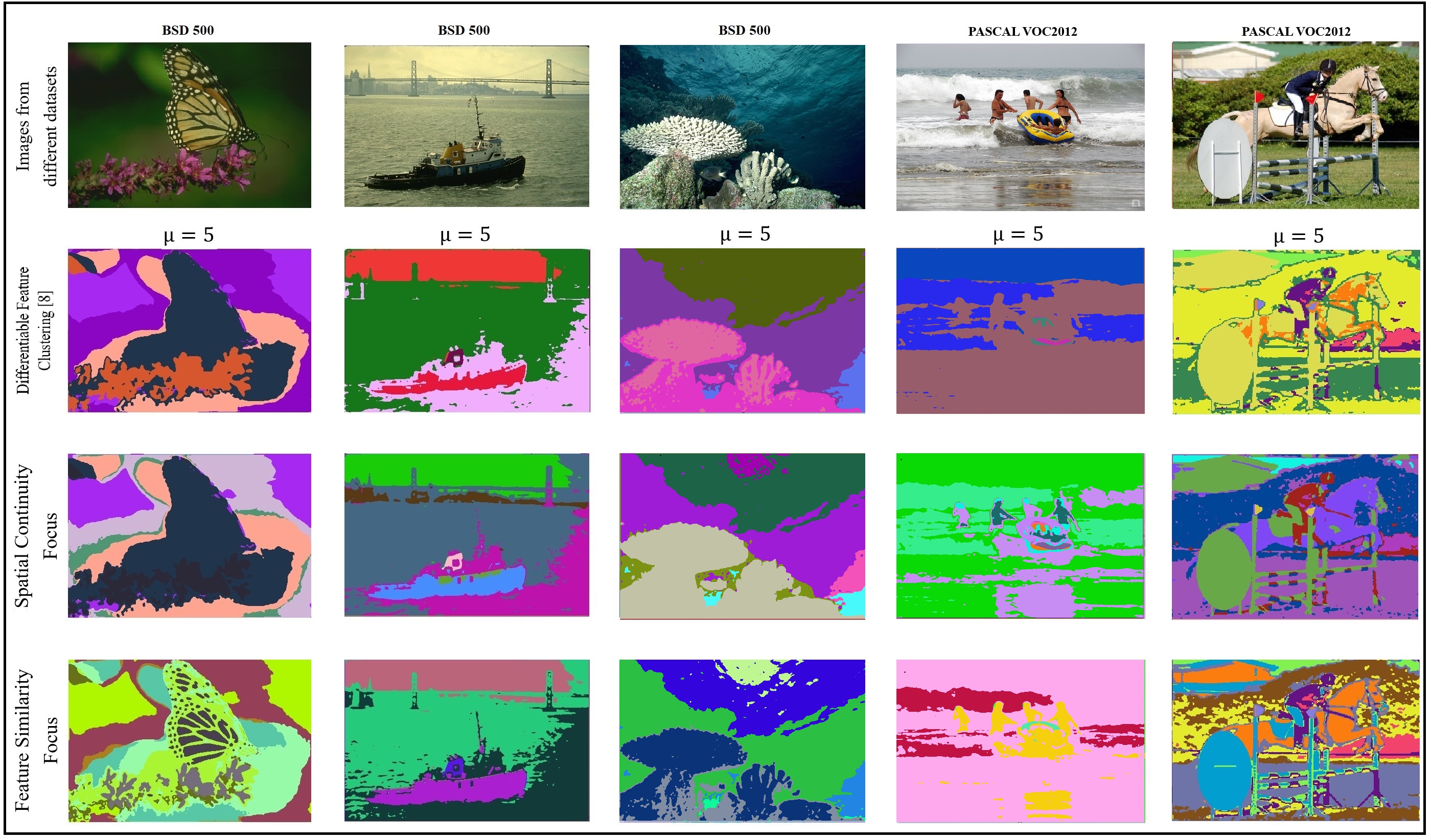}}
%  \vspace{1.5cm}
%\end{minipage}
%
\caption{Qualitative Results on select BSD500 and PASCAL VOC2012 images. Same color corresponds to the pixels being assigned the same clustering label by the algorithm. Please read Section \ref{qualres} for discussion on these results.}
\label{ALL2}
\end{figure*}
\subsection{Evaluation}

To assess the performance of the model trained without labels, a correspondence between the model's label space and the ground truth categories must be established. Initially, the model predicts on each image within the validation set. Subsequently, the confusion matrix is computed between the predicted labels and the ground truth classes. Employing linear assignment, we establish a one-to-one mapping between the predicted labels and ground truth classes, with the confusion matrix serving as the assignment cost. The mean Intersection over Union (mIoU) is then calculated over all classes based on this mapping \cite{Picie2021, DenseSiam}. For a more comprehensive analysis of the model's behavior, we present the mean IoU values for stuff and things classes.

Given the multiple ground truth types in BSD500, we adopt three mIOU counting strategies for assessment: "BSD500 All" considers all ground truth files, "BSD500 Fine" focuses on the ground truth file with the most significant number of segments, and "BSD500 Coarse" considers the ground truth file with the smallest number of segments. We define "BSD500 Mean" as the average of these three measurements.

\subsection{Quantitative Results}

The performance of various state-of-the-art methods in unsupervised semantic segmentation is comprehensively evaluated on multiple datasets.

\subsubsection{COCO-All Dataset}

Table \ref{tab:coco_all_results} presents the results on the COCO-All dataset, comparing mean Intersection over Union (mIoU) scores. Our proposed DynaSeg models, particularly the Feature Similarity Focus (FSF) and Spatial Continuity Focus (SCF) versions, exhibit significantly improved performance compared to existing methods. Notably, DynaSeg - SCF with ResNet-18 and FPN Extractor achieves the highest mIoU of 30.52, surpassing benchmarks like DenseSiam and Picie by substantial margins (14.12 and 16.08, respectively). This establishes our approach as the new state-of-the-art in unsupervised semantic segmentation.

\subsubsection{COCO-Stuff Dataset}

Table \ref{tab:coco_stuff_results} summarizes the results for the COCO-stuff dataset, considering different backbones and architectures. SCF with CNN-Based achieves an mIoU Stuff of 42.41 and pAcc of 76.7, demonstrating its effectiveness. However, FSF with ResNet-18 FPN emerges as the new state-of-the-art, surpassing all methods with the highest mIoU Stuff of 54.10 and pAcc of 81.1.
These results are particularly noteworthy because they surpass the performance of other models, including those utilizing Vision Transformer (ViT) architectures. For example, methods like STEGO \cite{hamilton2022unsupervised} and DINO + CAUSE-TR \cite{kim2023causal}, which are built on ViT-B/16 and ViT-B/8 backbones respectively, do not achieve the same level of performance as our DynaSeg models. Specifically, DINO + CAUSE-TR, which previously set a high benchmark with an mIoU Stuff of 41.9, is significantly outperformed by our DynaSeg - FSF model.

The exceptional performance of the ResNet-18 FPN integration in our model highlights its ability to enhance feature extraction through multi-resolution processing. This capability allows our model to capture details at different scales more effectively, which is crucial for achieving superior segmentation performance, especially on datasets with diverse object scales and complexities. This underscores the advantage of our approach in leveraging the strengths of both ResNet and FPN, even against advanced transformer-based models.

% \subsubsection{BSD500 and PASCAL VOC2012 Datasets}

% Quantitative results on BSD500 and PASCAL VOC2012 datasets are detailed in Table \ref{tab1}. SCF achieves a remarkable mIOU of 0.396 on PASCAL VOC2012, demonstrating its effectiveness. FSF consistently outperforms other methods on BSD500, emphasizing its robustness across various ground truth types. In addition to quantitative evaluations, we provide a demo \cite{DynaSeg_guermazi_video} of incremental segmentation using DynaSeg on a sample image from the Pascal VOC2012 dataset. This demo is available in the supplemental video, showcasing our method's application to real-world images.

\subsubsection{BSD500 and PASCAL VOC2012 Datasets}
The quantitative results on the BSD500 and PASCAL VOC2012 datasets are detailed in Table \ref{tab1}. DynaSeg's Spatial Continuity Focus (SCF) achieves a remarkable mIOU of 0.396 on PASCAL VOC2012, the highest among all methods compared, demonstrating its effectiveness in managing complex segmentation challenges. This performance underscores DynaSeg's adept handling of spatial details and continuity, which are crucial for achieving high-quality segmentation results. DynaSeg's Feature Similarity Focus (FSF) not only consistently outperforms other methods across the entire BSD500 dataset but also specifically surpasses the Graph-based Segmentation method on the BSD500 Fine dataset, showcasing its robustness and effectiveness in maintaining feature homogeneity across diverse segmentation tasks. This highlights FSF's distinct advantage over both traditional approaches like Graph-based Segmentation and recent innovations such as Differentiable Double Clustering with Edge-Aware Superpixel Fitting (DDCESF) \cite{li2024differentiable}, and other state-of-the-art techniques. In addition to quantitative evaluations, we provide a demo \cite{DynaSeg_guermazi_video} of incremental segmentation using DynaSeg on a sample image from the Pascal VOC2012 dataset. This demo, available in the supplemental video, effectively showcases the practical applications of DynaSeg to real-world images, illustrating its capabilities beyond traditional and contemporary unsupervised segmentation methods. 

In summary, the proposed FSF and SCF methods exhibit superior performance across different datasets and evaluation metrics. The achieved results reinforce their effectiveness, establishing them as prominent approaches in unsupervised semantic segmentation.

\subsection{Computational Efficiency Experiments}

To empirically demonstrate the computational efficiency of DynaSeg, we conducted experiments to compare the computational resources required by different segmentation methods. We measured the total parameters, and floating point operations per second (FLOPs) for each method. The results are presented in Table \ref{tab:computational_efficiency}.

\begin{table}[!t]
\centering
\begin{tabular}{|l|l|l|}
    \hline
    Method & Total Parameters & GFLOPs \\
    \hline
    DINO + HP \cite{seong2023leveraging} & 39,641,952 & 164.15 \\
    DINO\_Vit\_base\_16 \cite{caron2021emerging} & 86,415,592 & 16.86 \\
    IIC \cite{IIC} & 4,521,024 & 17.94 \\
    PiCIE \cite{Picie2021} & 23,631,424 & 4.32 \\
    DenseSiam \cite{DenseSiam} & 23,741,558 & 4.38  \\
    STEGO \cite{hamilton2022unsupervised} & 86,614,089 & 67.42 \\
    HSG \cite{ke2022unsupervised} & 27,968,704 & 7.13 \\
    DynaSeg (CNN\_based) & 193,900 & 9.75 \\
    DynaSeg (ResNet-18) & 12,046,272 & 1.84 \\
    \hline
\end{tabular}
\caption{Comparison of computational efficiency of different methods.}
\label{tab:computational_efficiency}
\end{table}

The results demonstrate that DynaSeg with both CNN-based and ResNet-18 backbones maintains competitive computational efficiency, with lower parameter counts and FLOPs compared to some state-of-the-art methods. Notably, the CNN-based version of DynaSeg demonstrates a good balance between computational cost and performance, making it suitable for applications requiring efficient processing. The results demonstrate that DynaSeg with ResNet-18 is highly efficient, requiring significantly fewer parameters and FLOPs compared to other state-of-the-art methods. For instance, DynaSeg (ResNet-18) uses only 1.84 GFLOPs and 12,046,272. This is significantly lower than methods such as DINO + HP \cite{seong2023leveraging}, which demands 164.15 GFLOPs and 39,641,952 parameters, and STEGO \cite{hamilton2022unsupervised}, which requires 67.42 GFLOPs and 86,614,089 parameters. The considerable reduction in computational requirements underscores DynaSeg's capability to deliver effective performance with much lower computational costs, making it particularly suitable for resource-constrained environments.
When examining the CNN-based version of DynaSeg, it maintains a competitive edge with only 193,900 parameters and 9.75 GFLOPs. Although methods like PiCIE \cite{Picie2021} and DenseSiam \cite{DenseSiam} show lower GFLOPs at 4.32 and 4.38 respectively, these efficiencies are achieved through the use of pre-trained models, which DynaSeg does not utilize. This distinction highlights the robustness and inherent efficiency of DynaSeg, as it achieves comparable or better performance without the need for extensive pre-training. The CNN-based DynaSeg, therefore, strikes an excellent balance between computational cost and segmentation performance, further validating its practical applicability.

In summary, the analysis reveals that DynaSeg, in both its ResNet-18 and CNN-based configurations, offers substantial computational advantages over other leading methods. The ResNet-18 version stands out for its minimal computational demands, positioning it as an optimal choice for various segmentation tasks where efficiency is paramount. Meanwhile, the CNN-based version provides a balanced approach, achieving efficiency and high performance independently of pre-trained models. These findings collectively affirm DynaSeg's effectiveness as a highly efficient and robust solution for unsupervised image segmentation, capable of meeting diverse application requirements while maintaining low computational overhead.

\subsection{Qualitative Results}
\label{qualres}

\begin{figure}[htb]

\hfill
\begin{minipage}[b]{1.0\linewidth}
  \centering
  \centerline{\includegraphics[width=9cm]{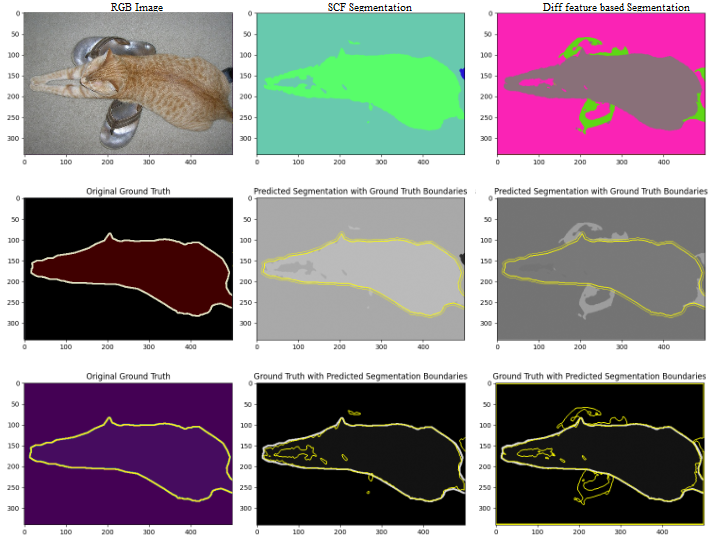}}
%  \vspace{1.5cm}
\end{minipage}
\caption{Qualitative results on Pascal VOC 2012: Original image, DynaSeg - SCF predicted segmentation, and Diff predicted segmentation.}
\label{fig:qualitative_results}
\end{figure}

We also provide qualitative results on a few images as done in \cite{them}. The qualitative evaluation showcases the segmentation results for the Spatial Continuity Focus (SCF) method, comparing it with the Differentiable Feature-based Segmentation model (Diff) \cite{them} and the ground truth. In Figure \ref{fig:qualitative_results}, SCF exhibits its strength in accurately capturing complex details of the cat, including fine contours, legs, and tail. The method achieves high-resolution segmentation with nuanced object boundaries, aligning effectively with the ground truth.

Notably, SCF outperforms Diff by providing a clean and accurate representation of the object. Diff, on the other hand, introduces noise in the background and struggles to accurately segment the tail of the cat, often blending it with the background. This emphasizes SCF's robust performance in preserving true object shapes and minimizing unwanted artifacts, making it particularly effective in complex scenes with detailed structures.

Additionally. As shown in Figure (\ref{ALL2}), our model is more effective in bringing out segmentation regions that are semantically related. For example, For the "Show Jumping" image (column 5), the horse and the obstacle are classified as the same class by \cite{them} (both yellow). However, for both FSF and SCF, the horse and the obstacle are appropriately distinguished. For the "ship" image (column 2), \cite{them} fails to differentiate between the sky and the body of the ship (both red), but both proposed SCF and FSF can do it successfully.

\begin{figure}[htb]

\hfill
\begin{minipage}[b]{1.0\linewidth}
  \centering
  \centerline{\includegraphics[width=8cm]{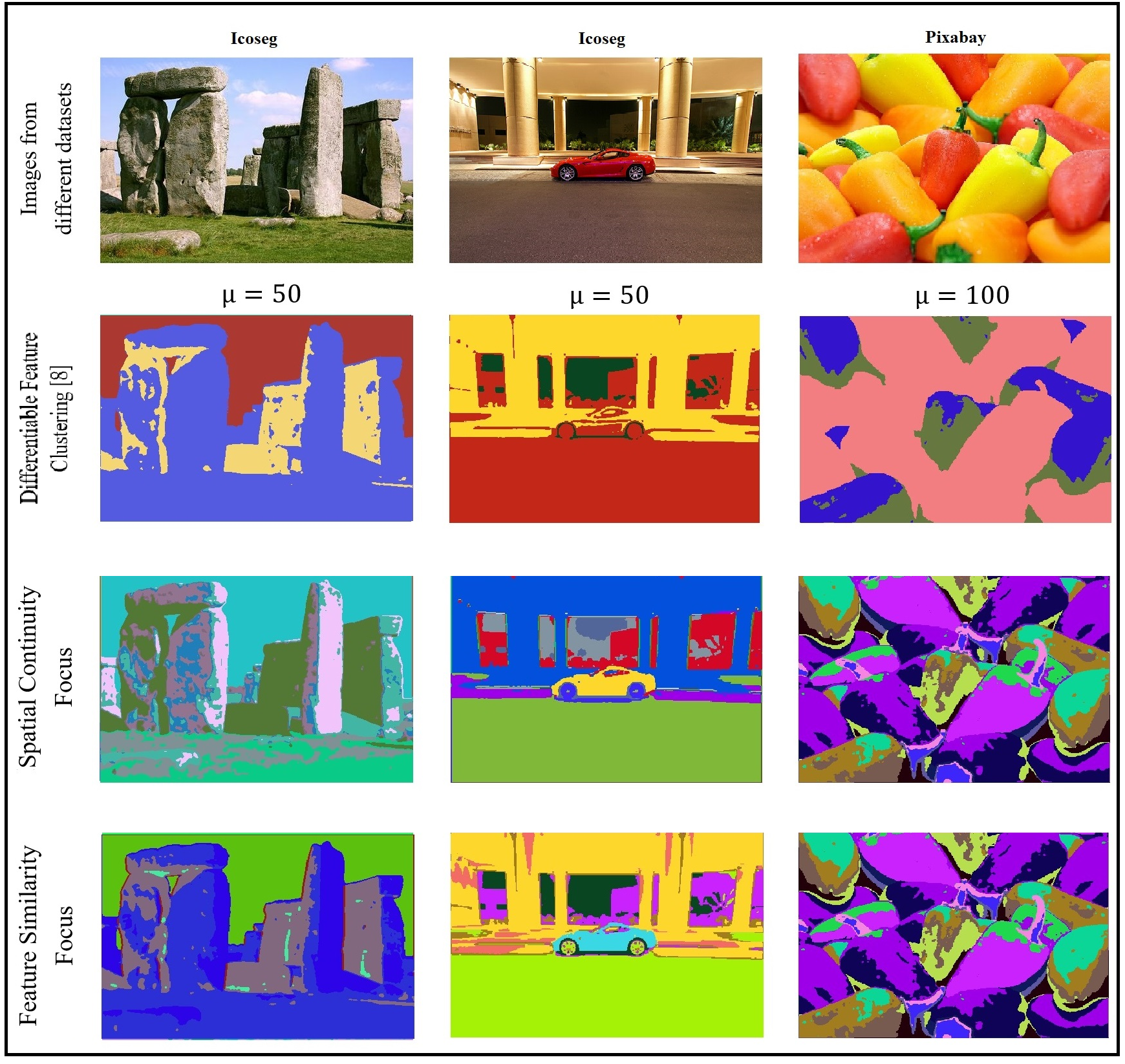}}
%  \vspace{1.5cm}
\end{minipage}
\caption{Qualitative Results on select Icoseg and Pixabay images. Same color corresponds to the pixels being assigned the same clustering label by the algorithm. Please read Section \ref{qualres} for discussion on these results.}
\label{ICOSEG_COMP2}

\end{figure}

Further qualitative results are shown for select Icoseg \cite{Icoseg} and Pixabay \cite{pixab} datasets that were also used in \cite{them}. These results can be seen in Figure (\ref{ICOSEG_COMP2}). The qualitative results on these datasets are presented to further demonstrate that our proposed approaches do not require as much parameter tuning as \cite{them} does. Figure (\ref{ICOSEG_COMP2}) highlights that the baseline Differentiable Feature Clustering \cite{them} is quite parameter sensitive. For each dataset, the weighting balance $\mu$ must be tuned extensively to obtain a more semantically meaningful result. For instance, PASCAL VOC 2012 and BSD500 datasets require a small balancing value $\mu=5$. While, Icoseg \cite{Icoseg} and Pixabay \cite{pixab} datasets need a much larger balance value $\mu=50$ and $\mu=100$, respectively. On the other hand, As illustrated in Figures (\ref{ALL2}) and (\ref{ICOSEG_COMP2}), our proposed method has proven effective in dealing with different datasets using the same weight for both FSF and SCF. The proposed methods also bring out details in an unsupervised manner that is semantically more meaningful. For example, using the Feature Similarity Focus method (FSF), the red car from the iCoseg dataset (column 2 row 4 in Figure (\ref{ICOSEG_COMP2})) displays more detail on the tires and more precise building outlines than the details extracted by \cite{them} (column 2 row 2), where the car is partially blended into the road. Similarly, for the peppers image (column 4 in Figure (\ref{ICOSEG_COMP2})), \cite{them} was unable to identify the shapes of the individual peppers accurately. Both of our proposed method do a much better job, even with the same value of $\mu$ as the other images. 

\begin{figure*}[htb]

\hfill
\begin{minipage}[b]{1.0\linewidth} 
  \centering
  \centerline{\includegraphics[width=18cm]{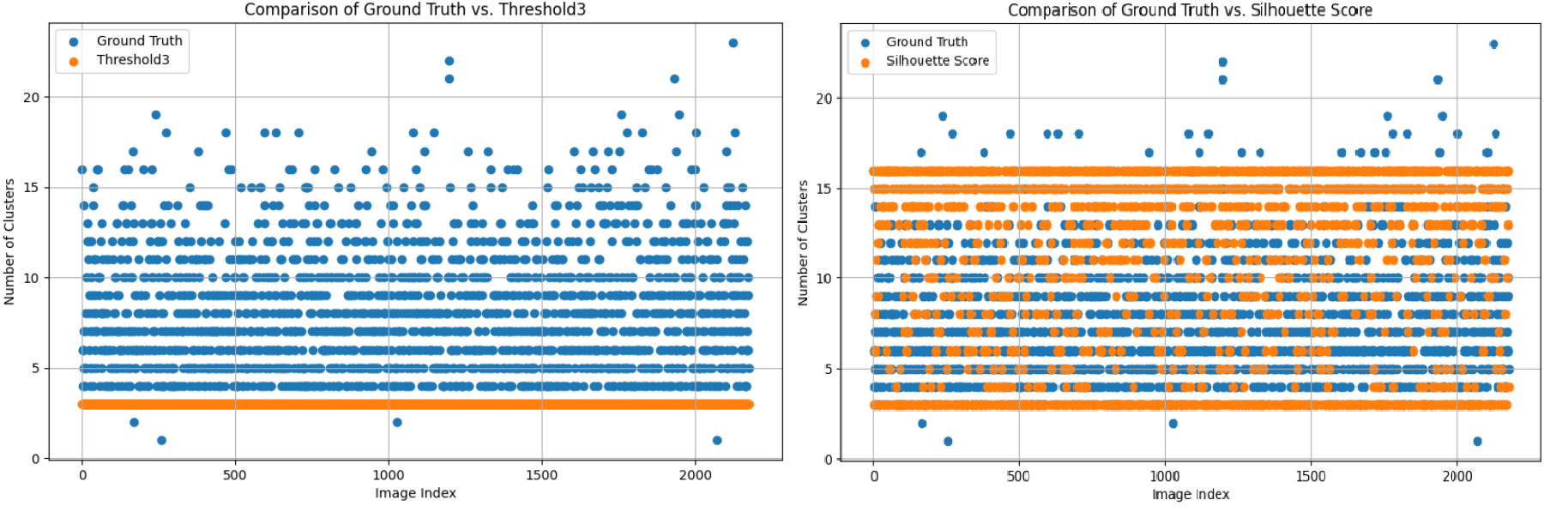}}
%  \vspace{1.5cm}
\end{minipage}
\caption{Comparison of cluster distribution: Left graph shows the comparison of the distribution of the number of clusters in the ground truth (blue) to a fixed threshold of three clusters (orange). Right graph displays the distribution of the number of clusters in the ground truth (blue) and the number of clusters predicted by the silhouette score (orange).}
\label{SilhouetteScore}
\end{figure*}
\begin{table}[!t]
\renewcommand{\arraystretch}{1.3}
\setlength{\tabcolsep}{3.5pt} % Adjust column spacing
% \extrarowheight
\centering
  \begin{tabular}{|c|c|c|c|c|c|}
    \hline
    Framework & Backbone & Thr & \multicolumn{3}{c|}{mIoU} \\
    \cline{4-6}
    &&& All & Things & Stuff \\
    \hline
    DynaSeg - FSF & CNN-Based & 3 & 26.03 & 61.73 & 34.82 \\
    DynaSeg - FSF & CNN-Based & Silh.S & {\textbf{30.51}} & 74.10 & {\textbf{42.39}} \\
    DynaSeg - SCF & ResNet-18 & 3 & \underline{30.52} & 74.06 & 42.41 \\
    DynaSeg - SCF & ResNet-18 & Silh.S & \underline{30.52} & 74.06 & 42.37 \\
    \hline
  \end{tabular}
  \caption{Silhouette Score Impact}
  \label{tab:Silhouette_Score_Impact}
\end{table}
In the comparative analysis of the proposed SCF and FSF methods, SCF exhibits notable proficiency in class segmentation, excelling in delineating distinct categories. Conversely, FSF demonstrates superior performance in instance segmentation tasks. Figure (\ref{ALL2}) column 3 illustrates this distinction. SCF accurately segments the image into well-defined classes such as water, the surface of the sea, and sea coral. On the other hand, FSF not only identifies these classes but also provides more details, capturing the nuances of sea coral structures, including the main trunk and branches. Notably, in Figure (\ref{ALL2}) column 1, FSF showcases its capability by revealing fine details of butterfly wing veins and forewing structures with remarkable precision. This precision makes FSF particularly well-suited for applications in the medical domain. The choice between SCF and FSF should be guided by the specific requirements of the given application, considering the emphasis on either class segmentation or detailed object representation.

The experimental results clearly demonstrate the superiority of our proposed dynamic weighting scheme over existing methods. The SCF and FSF methods consistently outperform other techniques in both quantitative and qualitative assessments. Specifically, the adaptability of the proposed approach to varying datasets and scenarios is evident in the significant improvements achieved in mIoU scores across different datasets and evaluation metrics.

\section{Ablation Study}
This section presents supplementary results examining the influence of the Silhouette Score Phase and the Feature Extractor through ablation on the COCO dataset. Additionally, we provide a comparison between FSF and SCF to further analyze their respective contributions. Furthermore, the integration of ResNet with the Feature Pyramid Network (FPN) significantly enhances feature extraction, leading to superior segmentation performance. This contribution highlights the effectiveness of our approach, particularly in handling objects at different scales and complexities.
\begin{table}[!t]
\renewcommand{\arraystretch}{1.3}
\setlength{\tabcolsep}{4pt}
\centering
  \begin{tabular}{|l|l|l|l|l|l|}
    \hline
    Framework & Backbone & \multicolumn{3}{c|}{mIoU} & pAcc\\
    \cline{3-5}
    && All & Things & Stuff& \\
    \hline
    DynaSeg - FSF & CNN-Based & {\bf30.51} & 74.10 & 42.39 & 76.65 \\
    DynaSeg - FSF & ResNet-18 &  30.07 & 62.87 & {\bf54.10} & {\bf81.08} \\
    DynaSeg - SCF & CNN-Based &  27.57 & 63.55 & 35.08 & 76.34 \\
    DynaSeg - SCF & ResNet-18  &  {\bf30.52} & 74.06 & \underline{42.37} & 76.67 \\
    \hline
  \end{tabular}
  \caption{Comparison of framework and backbone variations.}
  \label{tab:backbone_results}
\end{table}

\subsection{Silhouette Score Impact}

To evaluate the impact of integrating the Silhouette Score into our proposed dynamic weighting scheme, we conducted an ablation study, comparing segmentation results with and without the inclusion of Silhouette Score. Table \ref{tab:Silhouette_Score_Impact} presents the comparative results across different configurations.

In the case of the FSF method with a CNN-based backbone, integrating Silhouette Score at threshold 3 significantly improves the mIoU for "All" from 26.03 to \textbf{30.51} and for "Stuff" from 34.82 to \textbf{42.39}, demonstrating substantial enhancements.

For the SCF method with a ResNet-18 backbone, the segmentation results remain consistent for "All," "Things," and "Stuff" across both threshold 3 and Silh.S configurations, indicating that the model itself maintains the number of clusters, which doesn't reach the specified threshold. This implies that the SCF method inherently adjusts its clustering strategy, rendering the additional Silhouette Score less influential in this scenario. this nuanced understanding of the SCF behavior underscores the adaptability of the model and its ability to maintain segmentation quality without explicit reliance on external thresholds. It showcases the model's self-adjusting clustering mechanism, which may result in consistent mIoU values in certain scenarios.

The accompanying graph in Figure \ref{SilhouetteScore} provides additional insights into the distribution of ground truth cluster numbers, complementing the tabular results. The graph illustrates a significant disparity when comparing the distribution with a random threshold of 3. Conversely, when comparing the distribution with the optimal number of clusters derived from the Silhouette Score, a strong correlation is observed. This finding underscores the effectiveness of the proposed dynamic weighting scheme.

This combined analysis emphasizes the significance of incorporating the Silhouette Score in our framework, showcasing its positive influence on segmentation accuracy across different scenarios and backbone architectures.
\subsection{Effect of Feature Extractor}

To analyze the impact of the choice of feature extractor, we compare the performance of CNN-Based and ResNet-18 with FPN in the FSF and SCF methods. Table \ref{tab:backbone_results} comprehensively compares segmentation results across different combinations of frameworks and backbones. 

For the FSF method, the choice of the feature extractor has a substantial impact on segmentation performance. When using a CNN-Based backbone, the mIoU for "All" is 30.52, and for "Stuff," it is 42.39. In contrast, with a ResNet-18 backbone, the mIoU for "All" decreases slightly to 30.07, but there is a notable improvement in "Stuff" with a mIoU of 54.10. This suggests that the ResNet-18 backbone with FPN captures semantic information related to "Stuff" categories more effectively.

Similarly, for the SCF method, the impact of the feature extractor is evident. With a CNN-Based backbone, the mIoU for "All" is 27.57, and for "Stuff," it is 35.08. Switching to a ResNet-18 backbone results in an improvement, with the mIoU for "All" reaching 30.52 and for "Stuff" reaching 42.37. This highlights that the ResNet-18 backbone, even with its deeper architecture, contributes to better segmentation performance for both the "All" and "Stuff" categories.

The pixel accuracy (pAcc) values further complement these findings, showcasing the overall accuracy of pixel-level predictions. In summary, the choice of feature extractor, particularly the ResNet-18 backbone, plays a crucial role in enhancing segmentation performance, especially for specific categories like "Stuff."

\subsection{Comparison Between FSF and SCF}
To gain insights into the differences between the Feature Similarity Focus (FSF) and Spatial Continuity Focus (SCF) methods, we conduct a comparative analysis using Table \ref{tab:backbone_results}.

For the FSF method, with a CNN-Based backbone, the mIoU for "All" is 30.52, and for "Stuff," it is 42.39. On the other hand, the ResNet-18 backbone yields an mIoU of 30.07 for "All" and an impressive 54.10 for "Stuff." This indicates that FSF, particularly with a ResNet-18 backbone, excels in capturing semantic information related to complex and varied categories, such as "Stuff."

Moving to the SCF method, the mIoU for "All" with a CNN-Based backbone is 27.57, and for "Stuff," it is 35.08. Transitioning to a ResNet-18 backbone results in an improved mIoU of 30.52 for "All" and 42.37 for "Stuff." This suggests that SCF, like FSF, benefits from a more sophisticated backbone, and the ResNet-18 architecture contributes to better segmentation performance, especially for challenging categories like "Stuff."

To further refine the comparative analysis between the Feature Similarity Focus (FSF) and Spatial Continuity Focus (SCF) methods, we consider additional mIOU scores on BSD500 and PASCAL VOC2012 datasets, as presented in Table \ref{tab1}.
In the BSD500 dataset, across different segmentation scenarios (All, Fine, and Coarse), FSF consistently outperforms SCF in terms of mIOU scores. Notably, for BSD500 All, FSF achieves an mIOU of 0.349 compared to SCF's 0.330. This pattern is observed in other scenarios, reinforcing the effectiveness of FSF in capturing fine-grained details and semantic nuances.
On the PASCAL VOC2012 dataset, both FSF and SCF demonstrate competitive performance, with FSF achieving an mIOU of 0.391 and SCF's 0.396. The marginal difference in mIOU scores on PASCAL VOC2012 suggests that both methods perform comparably on this specific dataset, indicating their capability to handle diverse segmentation challenges effectively.

Comparing the two methods across different datasets and segmentation scenarios, FSF consistently delivers higher mIoU scores, indicating its robustness and effectiveness in capturing both global and fine-grained semantic information. The choice of FSF over SCF is particularly beneficial when targeting scenarios with varied and intricate segmentation requirements.
Both FSF and SCF perform better with a ResNet-18 backbone compared to CNN-based backbones. FSF, in particular, achieves higher mIoU values for "Stuff" across both backbones, emphasizing its effectiveness in capturing detailed and intricate features. Therefore, FSF shows promise for scenarios involving complex and diverse categories, especially when coupled with a ResNet-18 backbone.

\section{Conclusion}
% In conclusion, our study introduces a state-of-the-art unsupervised image segmentation approach that effectively overcomes the limitations of supervised methods, which heavily depend on costly pixel-level annotations. The dynamic weighting scheme, Silhouette Score Phase, integration of a pre-trained ResNet feature extraction, and Feature Pyramid Network (FPN) decoding collectively contribute to superior performance. Remarkably, our method achieves the pinnacle of segmentation accuracy without extensive parameter tuning, showcasing adaptability across diverse datasets. Our approach establishes itself as a state-of-the-art solution through extensive quantitative evaluation, demonstrating its efficacy in real-world applications.
Our study introduces a state-of-the-art unsupervised image segmentation approach. The dynamic weighting scheme, Silhouette Score Phase, and integration of a pre-trained ResNet feature extraction and Feature Pyramid Network (FPN) decoding collectively contribute to superior performance. The dynamic weighting scheme, in particular, enhances segmentation accuracy and flexibility by dynamically adjusting loss weights and cluster numbers during training, simplifying implementation and improving adaptability across diverse datasets. Additionally, the joint optimization framework and spatial continuity loss promote coherent and balanced segmentations by preserving high-frequency details and ensuring homogeneity within clusters. Our method achieves high segmentation accuracy without extensive parameter tuning, showcasing its adaptability across diverse datasets.

 \bibliographystyle{unsrt} 
 \bibliography{refs}

\end{document}